# Multi-View Learning in the Presence of View Disagreement

**C. Mario Christoudias**
UC Berkeley EECS & ICSI
MIT CSAIL

**Raquel Urtasun**
UC Berkeley EECS & ICSI
MIT CSAIL

**Trevor Darrell**
UC Berkeley EECS & ICSI
MIT CSAIL

## Abstract

Traditional multi-view learning approaches suffer in the presence of view disagreement, i.e., when samples in each view do not belong to the same class due to view corruption, occlusion or other noise processes. In this paper we present a multi-view learning approach that uses a conditional entropy criterion to detect view disagreement. Once detected, samples with view disagreement are filtered and standard multi-view learning methods can be successfully applied to the remaining samples. Experimental evaluation on synthetic and audio-visual databases demonstrates that the detection and filtering of view disagreement considerably increases the performance of traditional multi-view learning approaches.

## 1 Introduction

Many problems in machine learning involve datasets that are naturally comprised of multiple views, e.g, web pages can be classified from their content or the content of the pages that point to them, an object can be categorized from either its color or shape. In a multi-modal setting, multiple views can be defined from separate sensory modalities, e.g., a person's agreement can be classified from their speech utterance or head gesture. Approaches to *multi-view learning* [1, 3, 5, 7, 12, 15, 17, 22] exploit multiple redundant views to effectively learn from unlabeled data by mutually training a set of classifiers defined in each view[1]. Multi-view learning can be advantageous when compared to learning with only a single view [3, 4, 12], especially when the weaknesses of one view complement the strengths of the other.

A common assumption in multi-view learning is that the samples from each view always belong to the same class. In realistic settings, datasets are often corrupted by noise. Multi-view learning approaches have difficulty dealing with noisy observations, especially when each view is corrupted by an independent noise process. For example, in multi-sensory datasets it is common that an observation in one view is corrupted while the corresponding observations in other views remain unaffected (e.g., the sensor is temporarily in an erroneous condition before returning back to normal behavior). Indeed, if the corruption is severe, the class can no longer be reliably inferred from the corrupted sample.

These corrupted samples can be seen as belonging to a "neutral" or "background" class that co-occur with uncorrupted observations in other views. The view corruption problem is thus a source of *view disagreement*, i.e., the samples from each view do not always belong to the same class but sometimes belong to an additional background class as a result of view corruption or noise. In this paper we present a method for performing multi-view learning in the presence of view disagreement caused by view corruption. Our approach treats each view as corrupted by a structured noise process and detects view disagreement by exploiting the joint view statistics using a conditional entropy measure.

We are particularly interested in inferring multi-modal semantics from weakly supervised audio-visual speech and gesture data. In audio-visual problems view disagreement often arises as a result of temporary view occlusion, or uni-modal expression (e.g., when expressing agreement a person may say 'yes' without head nodding).

The underlying assumption of our approach is that *foreground* samples can co-occur with samples of the same class or background, whereas background samples can co-occur with samples from any class, a reasonable assumption for many audio-visual problems. We define new multi-view bootstrapping approaches that use conditional entropy in a pre-filtering step to reliably learn in the presence of view disagreement. Experimental evaluation on audio-visual data demonstrates that the detection and filtering of view disagreement enables multi-view learning to succeed despite large amounts of view disagreement.

[1] Note that the views are redundant in that each class can be inferred from both views separately. In the idealized setting each view would be conditionally independent given the class label (e.g., see [3]).

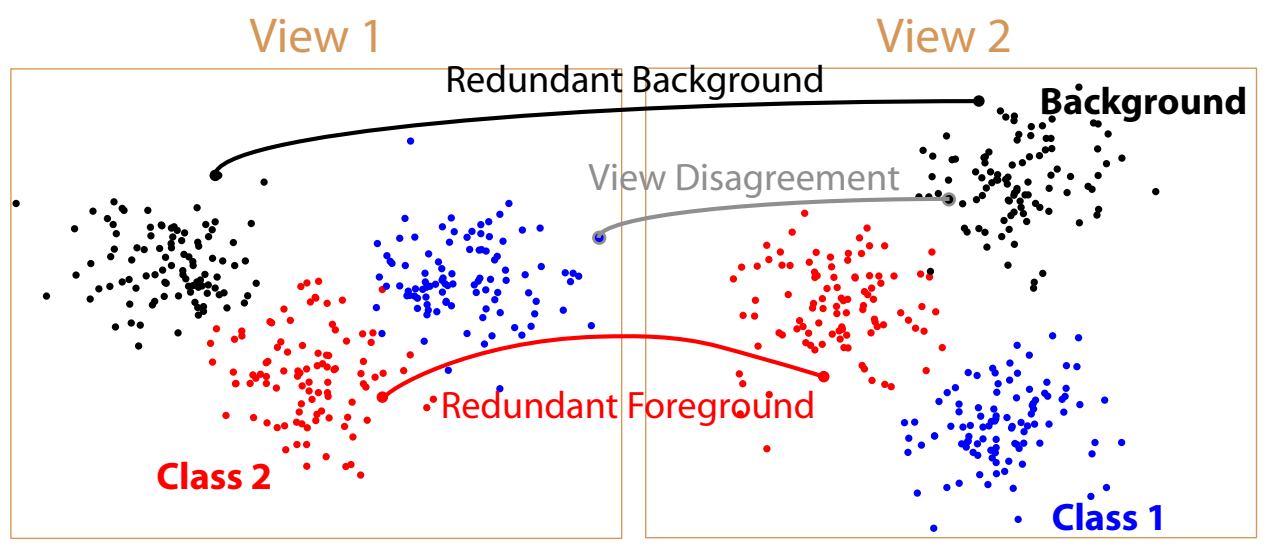

Figure 1: Synthetic two-view problem with normally distributed classes, two foreground and one background. Each view is 2-D; the two foreground classes are shown in red and blue. Corrupted samples form a separate background class (black samples) that co-occur with uncorrupted samples. For each point in the left view there is a corresponding point in the right view. Three point correspondences are shown: a redundant foreground sample, a redundant background sample and a sample with view disagreement where view 1 observed an instance of class 1, but view 2 for that sample was actually an observation of the background class. View disagreement occurs when one view is occluded and is incorrectly paired with background. Multi-view learning with these pairings leads to corrupted foreground class models.

The remainder of this paper is organized as follows. In the next section, a discussion of multi-view learning approaches and view disagreement is provided. Our conditional entropy based criterion for detecting view disagreement is then outlined in Section 3 and our multi-view bootstrapping approach is presented in Section 4. Experimental results are provided in Section 5. A discussion of related methods and connections between our work and other statistical techniques is given in Section 6. Finally, in Section 7 we provide a summary and discuss future work.

## 2 Multi-View Learning

Several approaches to multi-view learning have been proposed in the machine learning literature [1, 3, 5, 12, 15, 17, 22]. In their seminal work, Blum and Mitchell [3] introduced co-training which bootstraps a set of classifiers from high confidence labels. Nigam and Ghani [15] presented a co-EM algorithm that uses soft label assignment with EM to bootstrap classifiers from multiple views. Collins and Singer [5] proposed a co-boost approach that optimizes an objective that explicitly maximizes the agreement between each classifier. Similarly, Sindhwani et. al. [17] defined a co-regularization method that learns a multi-view classifier from partially labeled data using a view consensus-based regularization term. More recently, Yu et. al. [22] presented a Bayesian co-training framework that defines a multi-view kernel for semi-supervised learning with Gaussian Processes.

Although there exists a wide variety of multi-view learning algorithms, they all function on the common underlying principle of view agreement. More formally, let $\mathbf{x}_k = (x_k^1, ..., x_k^V)$ be a multi-view sample with $V$ views, and let $f_i : x^i \rightarrow \mathcal{Y}$ be the classifier that we seek in each view. Multi-view learning techniques train a set of classifiers $\{f_i\}$ by maximizing their consensus on the unlabeled data, $\mathbf{x}_k \in U$, for example by minimizing the $L_2$ norm [17],

$$\min \sum_{\mathbf{x_k} \in U} \sum_{i \neq j} \|f_i(x_k^i) - f_j(x_k^j)\|_2^2 \tag{1}$$

The minimization in Eq. (1) is only applicable to multi-view learning problems for which the views are *sufficient* for classification, i.e., that classification can be performed from either view alone. In practice, however, it is often difficult to define views that are fully sufficient. Previous methods for overcoming insufficiency have addressed the case where both views are necessary for classification [5, 2, 17]. These methods formulate multi-view learning as a global optimization problem that explicitly maximizes the consensus between views. Although these approaches allow for views with partial insufficiency, they still assume that each view is largely sufficient. In the presence of significant view disagreement these approaches would in general diverge and perform poorly.

In this paper we identify and address a new form of insufficiency inherent to many real-world datasets, caused by samples where each view potentially belongs to a different class, e.g., as a result of view corruption. We refer to this form of insufficiency as the *view disagreement* problem. The view disagreement problem is distinct from the forms of view insufficiency that have been addressed in the literature—previous methods for overcoming insufficiency have addressed the case where both views are necessary for classification [1, 5, 14, 17], but not the case where the samples from each view potentially belong to different classes.

The problem of view disagreement exists in many real-world datasets. In user agreement recognition from head gesture and speech [4], people often say 'yes' without head nodding and vice versa, and/or the subject can also become temporary occluded in either the audio or visual modalities by other speakers or objects in the scene. In semantic concept detection from text and video [21], it is possible for the text to describe a different event than what is being displayed in the video. Another example is web-page classification from page and hyper-link content [3], where the hyperlinks can often point to extraneous web-pages not relevant to the classification task.

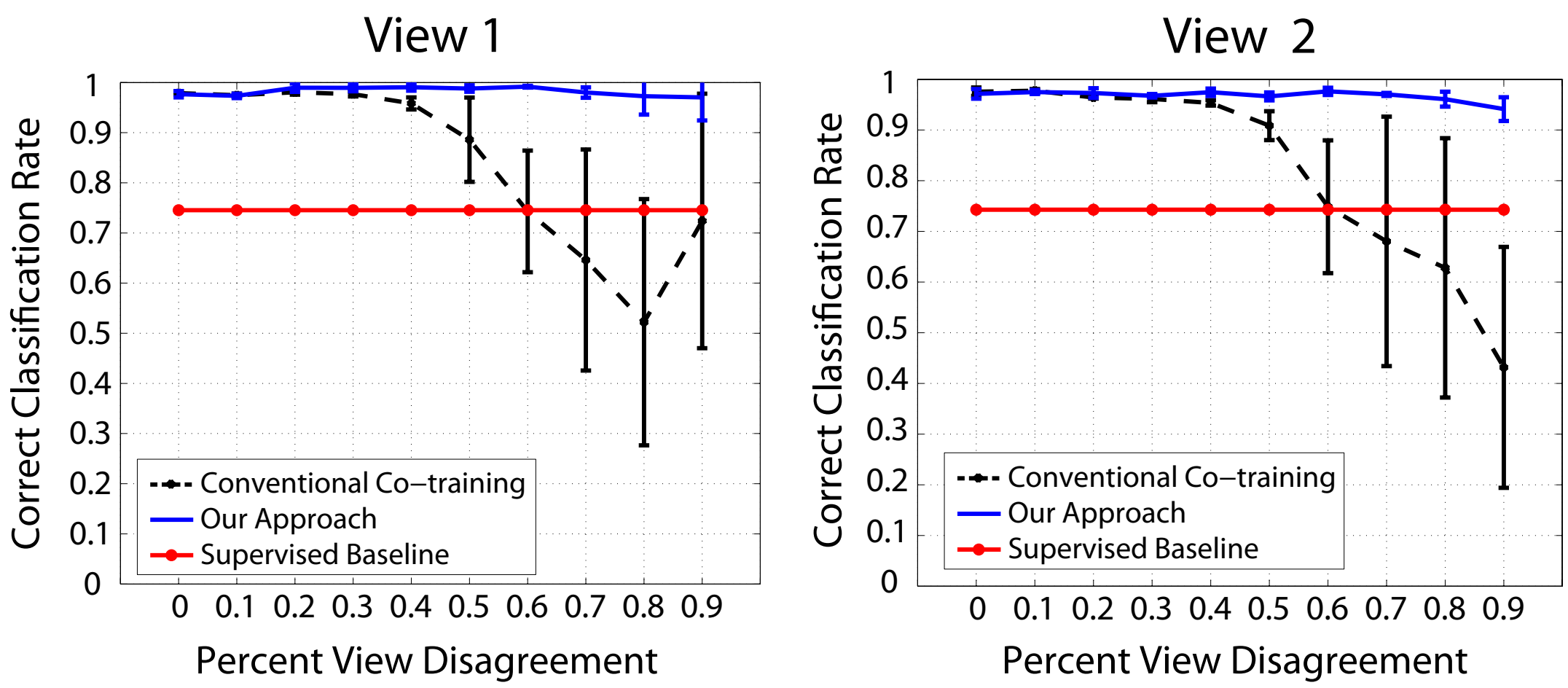

Figure 2: Multi-view learning for synthetic two-view example with varying amounts of view disagreement. Average performance is shown computed over 10 random splits of the training data into labeled and unlabeled sets; the error bars indicate ±1 std. devation. Our approach exhibits good performance at all view disagreement levels while conventional co-training begins to diverge for percent disagreement greater than 40%.

We illustrate the problem of view disagreement in multi-view learning with a toy example containing two views of two foreground classes and one background class. The samples of each class are drawn from Gaussian distributions with unit variance (see Figure 1). Figure 2 shows the degradation in performance of conventional co-training [3] for varying amounts of view disagreement. Here, co-training is evaluated using a left out test set and by randomly splitting the training set into labeled and unlabeled datasets. We report average performance across 10 random splits of the training data. As shown in Figure 2 co-training performs poorly when subject to significant amounts of view disagreement ($\geq 40\%$).

In what follows, we present a method for detecting view disagreement using a measure of conditional view entropy and demonstrate that when used as a pre-filtering step, our approach enables multi-view learning to succeed despite large amounts of view disagreement.

## 3 Detection and Filtering of View Disagreement

We consider an occlusion process where an additional class models background. We assume that this background class can co-occur with any of the $n+1$ classes in the other views[2], and that the $n$ foreground classes only co-occur with samples that belong to the same class or background, as is common in audio-visual datasets [4].

In this paper we propose a conditional entropy criterium for detecting samples with view disagreement. We further assume that background co-occurs with more than one foreground class; this is a reasonable assumption for many types of background (e.g., audio silence). In what follows, we treat each view $x^i$, $i = 1, ..., V$ as a random variable and detect view disagreement by examining the joint statistics of the different views. The entropy $H(x)$ of a random variable is a measure of its uncertainty [6]. Similarly, the conditional entropy $H(x|y)$ is a measure of the uncertainty in $x$ given that we have observed $y$. In the multi-view setting, the conditional entropy between views, $H(x^i|x^j)$, can be used as a measure of agreement that indicates whether the views of a sample belong to the same class or event. In what follows, we call $H(x^i|x^j)$ the *conditional view entropy*.

Under our assumptions we expect the conditional view entropy to be larger when conditioning on background

[2] Note that background samples can co-occur with the any of the $n$ foreground classes plus background.

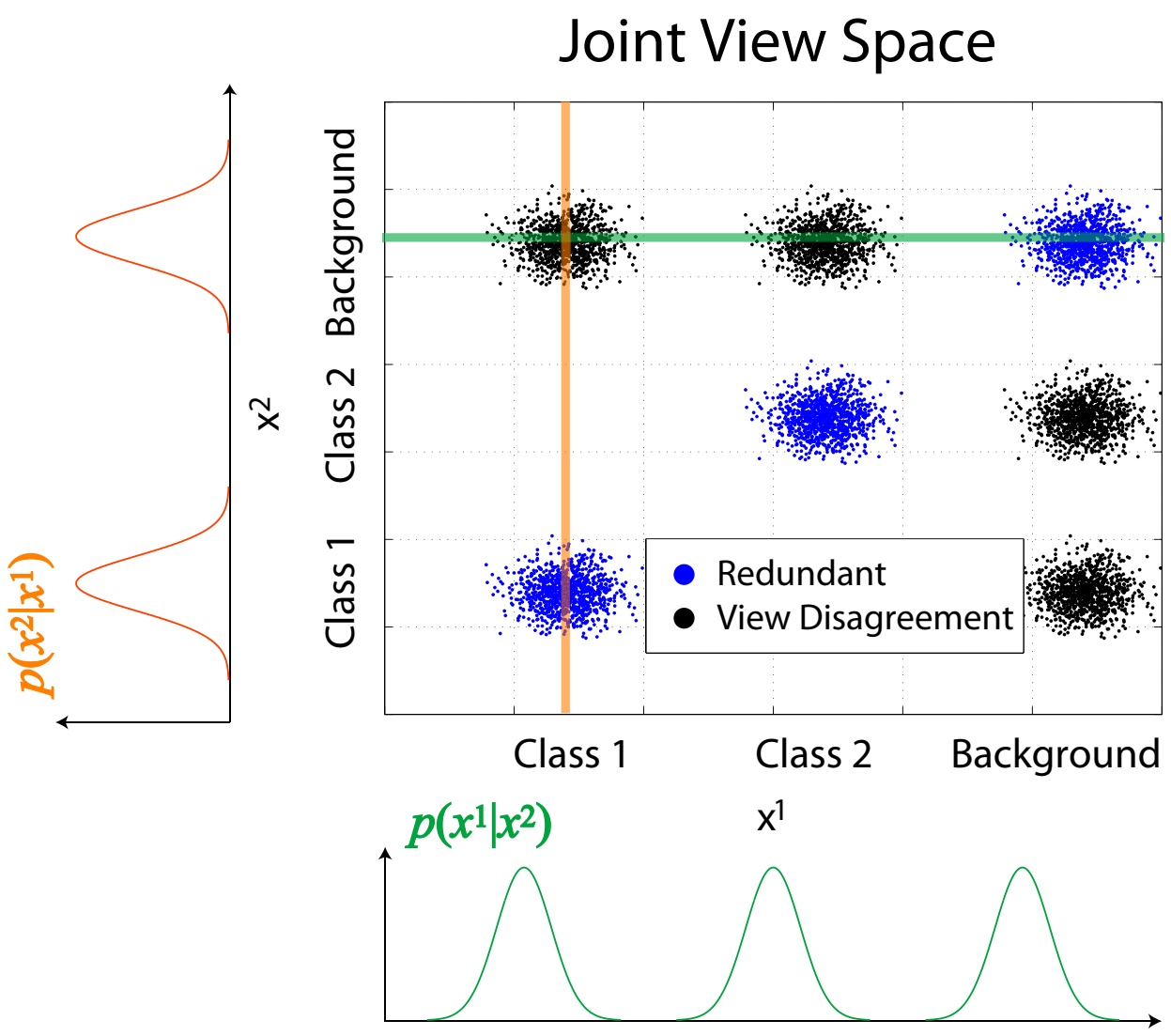

Figure 3: View disagreement caused by view corruption. The joint view space of a two-view problem with view disagreement is displayed. Redundant samples are highlighted in blue and samples with view disagreement in black. The conditional distributions for a sample with view disagreement are shown. The conditional distribution resulting from conditioning on background exhibits more peaks and therefore has a higher uncertainty than when conditioning on foreground.

compared to foreground. Thus, we have $\forall p = 1, ...n$,

$$H(x^i|x^j_k \in C^{n+1}) > H(x^i|x^j_l \in C^p) \qquad (2)$$

where $C^i$ is the set of examples belonging to class $i$. A notional example of view corruption is illustrated in Figure 3. This example contains two, 1-D views of two foreground classes and one background class. As before, the samples of each class are drawn from a normal distribution with unit variance. The conditional view distributions of a multi-view sample with view disagreement is displayed. Note that the uncertainty of view $i$ when conditioning on view $j$ has greater uncertainty when view $j$ is background.

We delineate foreground from background samples by thresholding the conditional view entropy. In particular, we define the threshold in each view using the mean conditional entropy computed over the unlabeled data. More formally, let $(x^i_k, x^j_k)$ be two different views of a multi-view sample $\mathbf{x}_k = (x^1_k, ..., x^V_k)$. We define an indicator function, $m(\cdot)$, that operates over view pairs $(x^i, x^j)$ and that is 1 if the conditional entropy of $x^i$ conditioned on $x^j_k$ is below the mean conditional entropy,

$$m(x^i, x^j_k) = \begin{cases} 1, & H(x^i|x^j_k) < \bar{H}_{ij} \\ 0, & \text{otherwise} \end{cases}, \qquad (3)$$

with

$$H(x^i|x^j_k) = -\sum_{x^i \in U^i} p(x^i|x^j_k) \log p(x^i|x^j_k), \qquad (4)$$

where $U^i$ is the $i$th view of the unlabeled dataset, and $\bar{H}_{ij}$ is the mean conditional entropy,

$$\bar{H}_{ij} = \frac{1}{M} \sum_{\mathbf{x}_k \in U} H(x^i|x^j_k), \qquad (5)$$

where $M$ is the number of samples in $U$. $m(x^i, x^j)$ is used to detect whether $x^j$ belongs to foreground since under our model foreground samples have a low conditional view entropy.

A sample $\mathbf{x}_k$ is a *redundant foreground* sample if it satisfies

$$\prod_{i=1}^{V} \prod_{j \neq i} m(x^i, x^j_k) = 1. \qquad (6)$$

Similarly, $\mathbf{x}_k$ is a *redundant background* sample if it satisfies

$$\sum_{i=1}^{V} \sum_{j \neq i} m(x^i, x^j_k) = 0. \qquad (7)$$

A multi-view sample $x_k$ is in *view disagreement* if it is neither a redundant foreground nor a redundant background sample.

**Definition 1.** Two views $(x^i_k, x^j_k)$ of a multi-view sample $\mathbf{x}_k$ are in view disagreement if

$$m(x^i, x^j_k) \oplus m(x^j, x^i_k) = 1 \qquad (8)$$

where $\oplus$ is the logical xor operator that has the property that $a \oplus b$ is 1 iff $a \neq b$ and 0 otherwise.

Eq. (8) defines our conditional entropy criterion for view disagreement detection between pairs of views of a multi-view sample.

In practice, we estimate the conditional probability of Eq. (4) as

$$p(x^i|x^j_k) = \frac{f(x^i, x^j_k)}{\sum_{x^i \in U^i} f(x^i, x^j_k)} \qquad (9)$$

where $f(\mathbf{x})$ is a multivariate kernel density estimator[3]. In our experiments, the bandwidth of $f$ is set using automatic bandwidth selection techniques [16].

[3] Note our approach is agnostic to the choice of probability model and more sophisticated conditional probability models can be used, such as [20], that perform better in high dimensional input spaces.

**Algorithm 1** Multi-View Bootstrapping in the Presence of View Disagreement

Given classifiers $f_i$ and labeled seed sets $S_i$, $i = 1, ..., V$, unlabeled dataset $U$ and parameters $N$ and $T$:
Set $t = 1$.
**repeat**
  **for** $i = 1, ..., V$ **do**
    Train $f_i$ on $S_i$
    Evaluate $f_i$ on $U^i$
    Sort $U$ in decreasing order by $f_i$ confidence
    **for** each $\mathbf{x}_k \in U$, $k = 1, ..., N$ **do**
      **for** $j \neq i$ **do**
        **if** $\neg(m(x^i, x_k^j) \oplus m(x^j, x_k^i))$ **then**
          $U^j = U^j \backslash \{x_k^j\}$
          $S^j = S^j \cup \{x_k^j\}$
        **end if**
      **end for**
      $U^i = U^i \backslash \{x_k^i\}$
      $S^i = S^i \cup \{x_k^i\}$
    **end for**
  **end for**
  Set $t = t + 1$.
**until** $|U| = \emptyset$ or $t = T$

## 4 Multi-view Bootstrapping in the Presence of View Disagreement

In this section we present a new multi-view bootstrapping algorithm that uses the conditional entropy measure of Eq. (8) in a pre-filtering step to learn from multi-view datasets with view disagreement.

Multi-view bootstrapping techniques, e.g., co-training, mutually train a set of classifiers, $f_i$, $i = 1, ..., V$, on an unlabeled dataset $U$ by iteratively evaluating each classifier and re-training from confidently classified samples. The classifiers are initialized from a small set of labeled examples typically referred to as the *seed set*, $S$. During bootstrapping, confidently classified samples in each view are used to label corresponding samples in the other views. It has been shown that multi-view bootstrapping is advantageous to self-training with only a single view [4].

We extend multi-view bootstrapping to function in the presence of view disagreement. A separate labeled set, $S_i$, is maintained for each view during bootstrapping and the conditional entropy measure of Eq. (8) is checked before labeling samples in the other views from labels in the current view. The parameters to the algorithm are $N$, the number of samples labeled by each classifier during each iteration of bootstrapping, and $T$ the maximum number of multi-view bootstrapping iterations. The resulting algorithm self-trains each classifier using all of the unlabeled examples, and only enforces a consensus on the samples with view agreement (see Algorithm 1).

**Algorithm 2** Cross-Modality Bootstrapping in the Presence of View Disagreement

Given existing classifier $f_1$ and initial classifier $f_2$, unlabeled dataset $U$ and parameter $N$:

*Initialization*:
Sort $U$ in decreasing order by $f_1$ confidence
Define $L = \{(y_k, x_k^2)\}$, $k = 1, ..., N$

*Bootstrapping*:
Set $S = \emptyset$
**for** each $(y_k, x^2) \in L$ **do**
  **if** $\neg(m(y, x_k^2) \oplus m(x^2, y_k))$ **then**
    $S = S \cup \{(y_k, x_k^2)\}$
    $L = L \backslash \{(y_k, x_k^2)\}$
  **end if**
**end for**
Train $f_2$ on $S$.

Figure 2 displays the result of multi-view bootstrapping for the toy example of Figure 1 using $N = 6$ and $T$ was set such that all the unlabeled data was used. With our method, multi-view learning is able to proceed successfully despite the presence of severe view disagreement and is able to learn accurate classifiers in each view even when presented with datasets that contain up-to 90% view disagreement.

In audio-visual problems it is commonly the case that there is an imbalance between the classification difficulty in each view. In such cases, an accurate classifier can be learned in the weaker view using an unsupervised learning method that bootstraps from labels output by the classifier in the other view. Here, the class labels output by the classifier in the stronger view can be used as input to the conditional entropy measure as they provide a more structured input than the original input signal.

The resulting cross-modality bootstrapping algorithm trains a classifier $f_2$ in the second view from an existing classifier $f_1$ in the first view on a two-view unlabeled dataset $U$. The algorithm proceeds as follows. First $f_1$ is evaluated on $U$ and the $N$ most confidently classified examples are moved from $U$ to the labeled set $L$. The conditional entropy measure is then evaluated over each label, sample pair $(y, x^2) \in L$, where $y = f_1(x^1)$. The final classifier $f_2$ results from training on the the samples in $L$ that are detected as redundant foreground or redundant background (see Algorithm 2).

## 5 Experimental Evaluation

We evaluate the performance of multi-view bootstrapping techniques on the task of audio-visual user agreement recognition from speech and head gesture. Al-though users often use redundant expression of agreement, it is frequently the case that they say ‘yes’ without head gesturing and viceversa. View disagreement can also be caused by noisy acoustic environments (e.g., a crowded room), temporary visual occlusions by other objects in the scene, or if the subject is temporarily out of the camera’s field of view.

To evaluate our approach we used a dataset of 15 subjects interacting with an avatar in a conversational dialog task [4]. The interactions included portions where each subject answered a set of yes/no questions using head gesture and speech. The head gesture consisted of head nods and shakes and the speech data of ‘yes’ and ‘no’ utterances. In our experiments, we simulate view disagreement in the visual domain using both no motion (i.e., random noise) and real background head motion samples from non-response portions of the interaction. Similarly, background in the audio is simulated as babble noise.

The visual features consist of 3-D head rotation velocities output by a 6-D head tracker [13]. For each subject, we post-process these observations by computing a 32 sample windowed Fast Fourier Transform (FFT) separately over each dimension, with a time window of 1 second corresponding to the expected length of a head gesture. The resulting sequence of FFT observations is then segmented using the avatar’s transcript which marks the beginning and end of each user response.

The FFT spectra of each user response were amplitude normalized and blurred in space and time to remove variability to location, duration and rate of head motion. Principle Components Analysis (PCA) was then performed over the vector space resulting from flattening the FFT spectra corresponding to each response into a single vector. The resulting 3-D PCA space captured over 90% of the variance and was computed over the unlabeled samples of the training set.

The audio features consist of 13-D Mel Frequency Cepstral Coefficients (MFCCs) sampled at 100Hz over the segmented audio sequences corresponding to each user response, obtained from the avatar’s transcript. The audio sequences were then converted into single frame observations using the technique of [11]. In this representation, an audio sequence is divided into portions and an average MFCC vector is computed over each portion. In our experiments, we used proportions equal to $(0.3, 0.4, 0.3)$. The concatenated averages along with first derivatives and log duration define a 61-D observation vector. To reduce the dimensionality of this space, PCA was applied retaining 98% of the variance which resulted in a 9-D, single-frame audio observation space.

In our experiments we use correct classification rate as the evaluation metric, defined as:

$$\text{CCR} = \frac{\#\text{ of examples correctly classified}}{\text{total }\#\text{ of examples}} \qquad (10)$$

We used Bayes classifiers for audio and visual gesture recognition defined as $p(y|x) = \frac{p(x|y)}{\sum_y p(x|y)}$, where $p(x|y)$ is Gaussian. Specifically, Bayes classifiers for $p(y|x^a)$ and $p(y|x^v)$ are bootstrapped from semi-supervised audio-visual data; $x^a$ and $x^v$ correspond to audio and visual observations respectively.

### 5.1 Cross-Modality Bootstrapping

First, we evaluate our cross-modality bootstrapping approach. For this task, we are interested in performing semi-supervised learning of visual head gesture by bootstrapping from labels in the speech (e.g., those output by an off-the-shelf speech recognizer). We simulated view disagreement by randomly replacing observations in the visual modality with background sequences, and replacing labels in the audio with the background label. Redundant background was also added such that there were an equal number of redundant background samples as there were redundant foreground samples per class.

We first show results using a “no motion” visual background modeled as zero mean Gaussian noise in the 3-D head rotational velocity space with $\sigma = 0.1$. Figure 4 displays the result of evaluating the performance of multi-view bootstrapping (Algorithm 2) with varying amounts of view disagreement. Performance is shown averaged over 5 random splits of the data into 10 train and 5 test subjects. At small amounts of view disagreement ($\leq$ 20%) conventional bootstrapping and our approach exhibit similar good performance. When the view disagreement is small the error can be viewed as classification noise in the audio. For larger amounts of view disagreement (up to 50%), conventional multi-view bootstrapping diverges and our algorithm still succeeds in learning an accurate head gesture recognizer from the audio-visual data. For $>$ 50% view disagreement, our approach begins to degrade and exhibits a large variance in performance. This high variability can be a result of poor bandwidth selection, or a poor choice of threshold. We plan to investigate alternative methods for modeling the conditional probability and more sophisticated threshold selection techniques as part of future work.

Figure 4(b) displays average receiver-operator curves (ROCs) for redundant foreground and background class detection that result from varying the entropy threshold of the conditional entropy measure. The mean conditional entropy defines a point on these curves. As illustrated by the figure, overall our approach does fairly well in detecting view disagreement.

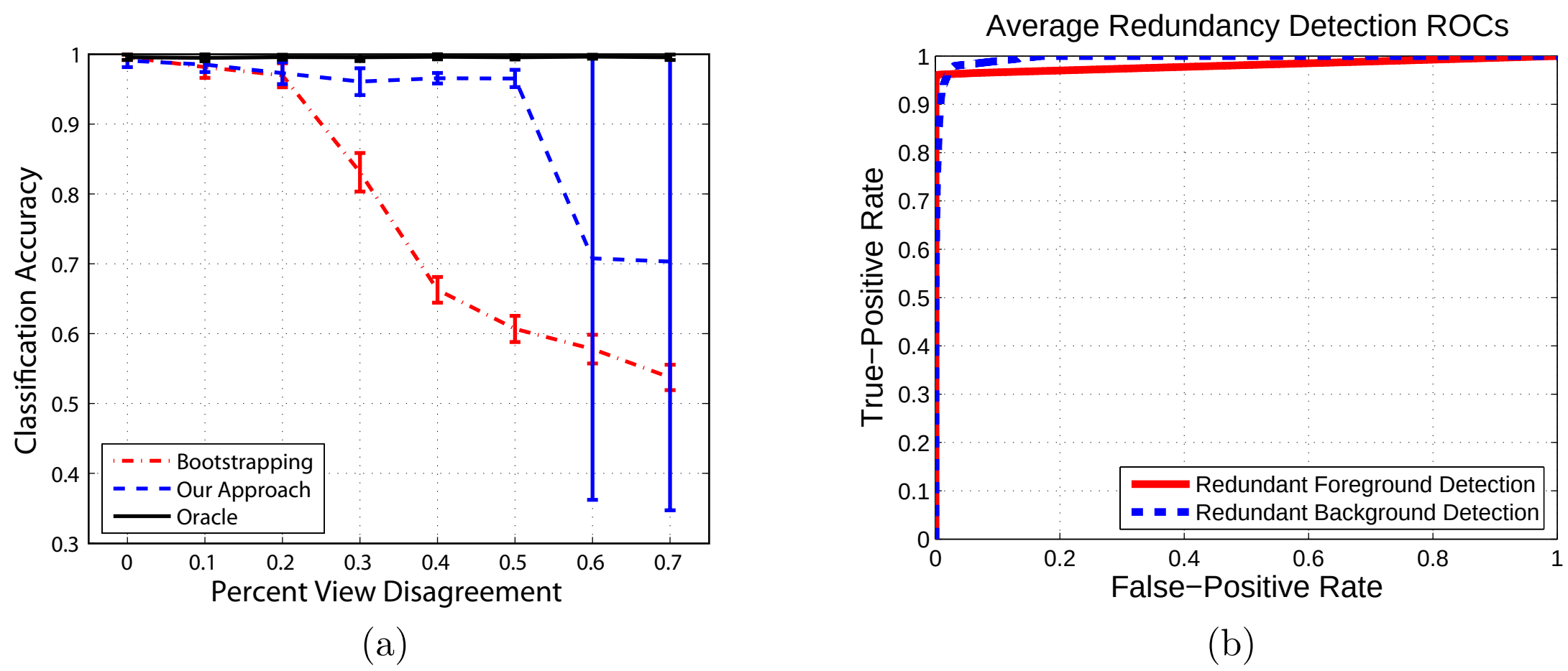

Figure 4: Bootstrapping a user agreement visual classifier from audio. (a) Performance is shown averaged over random splits of the data into 10 train and 5 test subjects over varying amounts of simulated view disagreement using a no motion background class; error bars indicate ± 1 std. deviation. Unlike conventional bootstrapping, our approach is able to cope with up-to 50% view disagreement. (b) Average view disagreement detection ROCs are also shown for redundant foreground and background detection. Our approach effectively detects view disagreement.

Next, we consider a more realistic occluding visual background class generated from randomly selecting head motion sequences from non-response portions of each user interaction. In contrast to the “no motion” class considered above, these segments contain miscellaneous head motion in addition to no motion.

Our view disagreement detection approach (Algorithm 2) performs equally well in the presence of the more challenging real background as is shown in Figure 5. As before, conventional bootstrapping performs poorly in the presence of view disagreement. In contrast, our approach is able to successfully learn a visual classifier in the presence of significant view disagreement (up to 50%).

### 5.2 Multi-View Bootstrapping

We evaluated the performance of multi-view bootstrapping (Algorithm 1) for the task of semi-supervised learning of audio-visual user agreement classifiers from speech and head gesture. Figure 6 displays the result of audio-visual co-training for varying amounts of view disagreement. Performance is shown averaged over 5 random splits of the data into 10 train and 5 test subjects and over 10 random splits of the training data into labeled seed set and unlabeled training set, with 15 labeled samples, 5 per class. Conventional co-training and our approach were then evaluated using $N = 6$ and $T = 100$. We chose $N$ such that the classes are balanced.

For this problem, the initial visual classifier trained from the seed set is much more accurate than the initial audio classifier that performs near chance. The goal of co-training is to learn accurate classifiers in both the audio and visual modalities. Note, that in contrast to cross-modality bootstrapping, this is done without any a priori knowledge as to which modality is more reliable. For small amounts of view disagreement ($\geq 20\%$), both conventional co-training and our approach (Algorithm 1) are able to exploit the strong performance in the visual modality to train an accurate classifier in the audio. For larger amounts of view disagreement, conventional co-training begins to diverge and at the 70% view disagreement level is not able to improve over the supervised baseline in both the audio and visual modalities. In contrast, our approach reliably learns accurate audio-visual classifiers across all view disagreement levels.

## 6 Discussion

Recently, Ando and Zhang [1] presented a multi-view learning approach that instead of assuming a consensus over classification functions assume that the views share the same low dimensional manifold. This has the advantage that it can cope with insufficient views where classification cannot be performed from either view alone. Still, their approach defines a consensus between views, and therefore assumes that the samples in each view are of the same class. View disagreement will violate this assumption and we expect their method to degrade as multi-view bootstrapping.

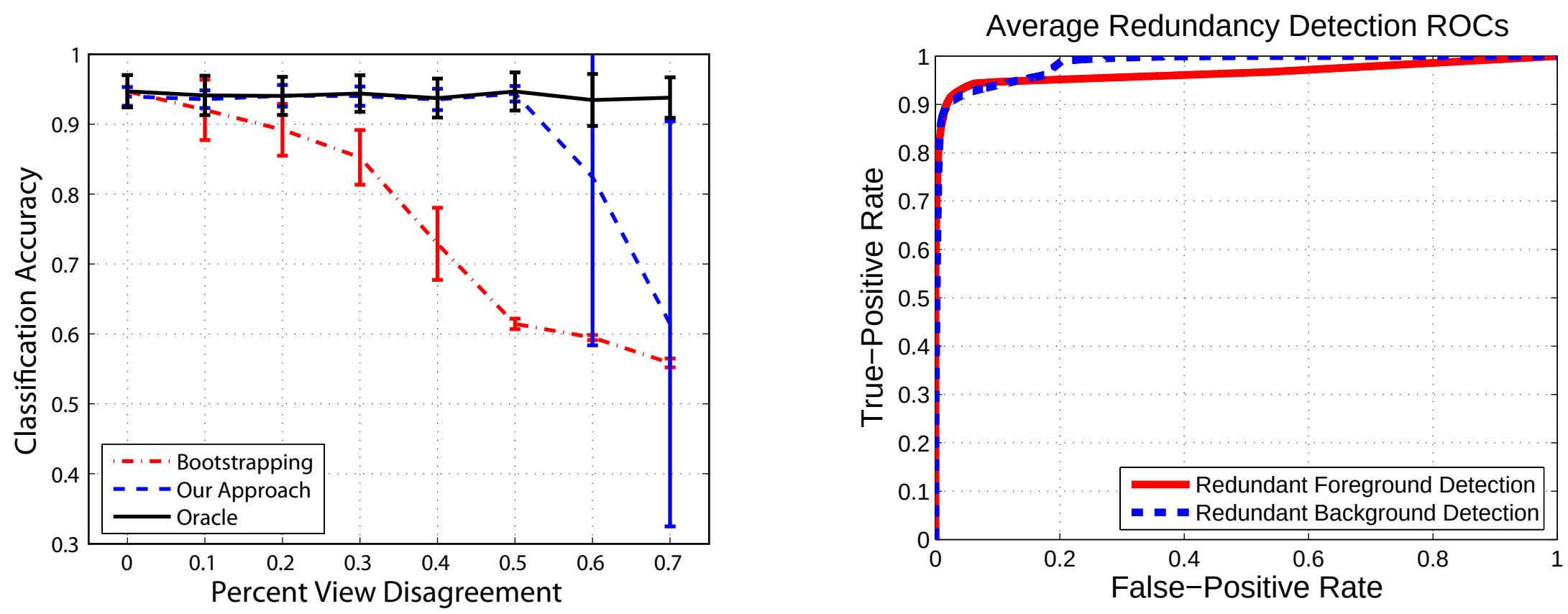

(a) (b)

Figure 5: Bootstrapping a user agreement visual classifier from audio with real visual background. Performance is shown averaged over random splits of the data into 10 train and 5 test subjects over varying amounts of simulated view disagreement; error bars indicate ± 1 std. deviation. The conventional bootstrapping baseline performs poorly in the presence of view disagreement. In contrast, our approach is able to (a) successfully learn a visual classifier and (b) classify views in the presence of significant view disagreement (up to 50%).

Our approach treats each view as corrupted by a structured noise process and detects view disagreement by exploiting the joint view statistics. An alternative method to coping with view disagreement is to treat each view as belonging to a stochastic process and use a measure such as mutual information to test for view dependency [19, 18]. In [18], Siracusa and Fisher use hypothesis testing with a hidden factorization Markov model to infer dependency between audio-visual streams. It would be interesting to apply such techniques for performing multi-view learning despite view disagreement, which we leave as part of future work.

Our work bears similarity to co-clustering approaches which use co-occurrence statistics to perform multi-view clustering [10, 9, 8]. These techniques, however, do not explore the relationship between co-occurrence and view sufficiency and would suffer in the presence of view disagreement since the occluding background would potentially cause foreground clusters to collapse into a single cluster.

We demonstrated our view disagreement detection and filtering approach for multi-view bootstrapping techniques (e.g., [3, 15, 4]). However, our algorithm is generally applicable to any multi-view learning method and we believe it will be straightforward to adapt it for use with other approaches (e.g., [1, 5, 17]). Multi-view learning methods either implicity or explicitly maximize the consensus between views to learn from unlabeled data; view disagreement adversatively affects multi-view learning techniques since they encourage agreement between views.

In our experiments, our approach performs well on a realistic dataset with noisy observations. The success of our approach on this dataset is predicated on the fact that foreground and background classes exhibit distinct co-occurrence patterns, which our algorithm exploits to reliably detect view disagreement.

## 7 Conclusions and Future Work

In this paper we have identified a new multi-view learning problem, view disagreement, inherent to many real-word multi-view datasets. We presented a multi-view learning framework for performing semi-supervised learning from multi-view datasets in the presence of view disagreement and demonstrated that a conditional entropy criterion was able to detect view disagreement caused by view corruption or noise. As shown in our experiments, for the task of audio-visual user agreement our method was able to successfully perform multi-view learning even in the presence of gross view disagreement ($50-70\%$). Interesting avenues for future work include the investigation of alternative entropy threshold selection techniques, the use of alternative probability models for computing conditional entropy and modeling redundancy between non-stationary stochastic processes using measures such as mutual information.

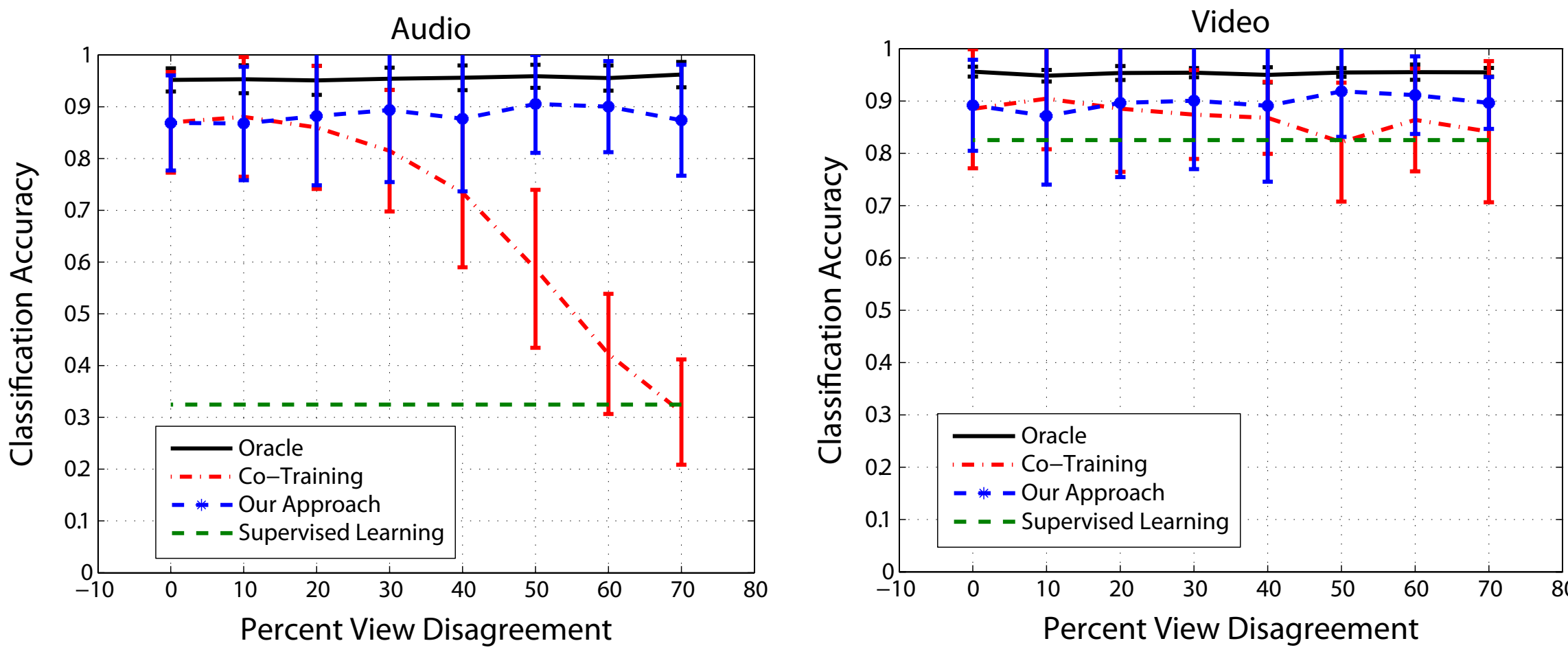

(a) (b)

Figure 6: Multi-view bootstrapping of audio-visual user agreement classifiers. Performance of (a) audio and (b) video is displayed averaged over 5 random splits of the data into 10 train and 5 test subjects and over 10 random splits of each training set into labeled seed set and unlabeled dataset; error bars show $\pm 1$ std. deviation. Conventional co-training performs poorly in the presence of significant view disagreement. In contrast, our approach performs well across all view disagreement levels.